\documentclass[11pt,a4paper]{article}
\usepackage[utf8]{inputenc}
\usepackage[english]{babel}
\usepackage{mathptmx}
\usepackage[margin=2.5cm]{geometry}
\usepackage{amsmath}
\usepackage{graphicx}
\usepackage{adjustbox}
\usepackage{textcomp}
\usepackage{multirow}
\usepackage{booktabs}
\usepackage{microtype}
\usepackage[labelfont=bf]{caption}
\usepackage[labelfont=bf]{subcaption}
\usepackage{xcolor}
\usepackage{tikz}
\usetikzlibrary{positioning,arrows.meta,fit,shapes.geometric,decorations.pathreplacing,calc}
\usepackage[numbers,sort&compress]{natbib}
\usepackage{authblk}

\usepackage{url}
\usepackage[colorlinks,linkcolor=blue!50!black,citecolor=blue!50!black,urlcolor=blue!50!black,breaklinks]{hyperref}
\usepackage{orcidlink}
\definecolor{color1}{RGB}{0,0,90}
\definecolor{color2}{RGB}{0,20,20}
\hypersetup{pdftitle={Protocol Choices Matter More Than the Aggregation Algorithm: Federated Learning for In-Hospital Mortality Prediction in Care-Unit and Multi-Hospital Federations},pdfauthor={Rodrigo Tertulino, La\'ercio Alencar, Ricardo Almeida}}

\title{Protocol Choices Matter More Than the Aggregation Algorithm: Federated Learning for In-Hospital Mortality Prediction in Care-Unit and Multi-Hospital Federations}

\author[1,*]{Rodrigo Tertulino\,\orcidlink{0000-0002-7594-9312}}
\author[1]{La\'ercio Alencar\,\orcidlink{0009-0007-5226-9019}}
\author[1]{Ricardo Almeida\,\orcidlink{0009-0006-3447-7431}}
\affil[1]{Instituto Federal de Educa\c{c}\~ao, Ci\^encia e Tecnologia do Rio Grande do Norte (IFRN), Brazil\protect\\
\texttt{rodrigo.tertulino@ifrn.edu.br}, \texttt{lapea@ifrn.edu.br}, \texttt{ricardo.almeida@ifrn.edu.br}}
\affil[*]{Corresponding author: \texttt{rodrigo.tertulino@ifrn.edu.br}}
\date{}

\begin{document}
\maketitle

\begin{abstract}
Federated learning allows hospitals to train a shared mortality model without pooling their records, but published comparisons of aggregation strategies rarely control the evaluation protocol. This study compared seven strategies (FedAvg, FedProx, SCAFFOLD, FedAdagrad, FedAdam, FedYogi and FedCluster) on two federations built with one pipeline: 12 care units of a hospital in MIMIC-IV and 49 hospitals in eICU-CRD. Each strategy was tuned and selected on validation data by the same rule, and every comparison was repeated across 10 seeds with Friedman and Holm-corrected tests. SCAFFOLD ranked first in both federations (AUC-ROC 0.886 and 0.819) and in 17 of 18 variants, but its margin over the runner-up never exceeded 0.008, and a control sharing its local optimizer attributed most of its margin over FedAvg to drift correction. On the multi-hospital federation, federated training matched centralized training. In the care-unit federation, three protocol choices changed discrimination by more than the choice of strategy did: discharge-coded predictors (+0.038 in AUC-ROC), client-level SMOTE (up to \textminus0.074), and evaluating the last instead of the best validation round (up to \textminus0.084). Platt scaling on validation data repaired the risks distorted by imbalance corrections without loss of discrimination. Studies should therefore fix the protocol before ranking algorithms.

\end{abstract}

\section{Introduction}
\label{sec:intro}
Early prediction of in-hospital mortality from electronic health records (EHR) is one of the most studied tasks in clinical machine learning, both because it supports triage and resource allocation and because it serves as a standard benchmark for new methods~\cite{harutyunyan2019}. Models of this kind improve with the amount and diversity of the data they learn from, yet the records that would provide that diversity are spread across hospitals that cannot pool them. Privacy regulation, institutional policy, and the cost of building central repositories all stand in the way. Federated learning addresses this obstacle by training a shared model while the records stay where they were collected: each institution trains on its own data, and only model updates are sent to a server that aggregates them~\cite{mcmahan2017,rieke2020}.

How the server aggregates the updates is the central design choice of a federated system, and a family of aggregation strategies has been proposed for the situation that hospitals typify, in which clients differ in size and case mix~\cite{kairouz2021}. Practitioners who must choose among these strategies depend on empirical comparisons on clinical data, but the available comparisons were not designed to isolate the effect of the strategy. The most comprehensive comparison on EHR mortality prediction found small differences between six strategies on a federation of hospitals~\cite{dang2022}; it used shared hyperparameters whose values were not reported, took the federated models at the last round but the centralized model at its best validation epoch, and evaluated a single run of each method. Other common practices add further uncertainty: oversampling the rare outcome inside each client distorts the predicted risks~\cite{goorbergh2022}, and predictors that are recorded only after the prediction time inflate performance~\cite{kapoor2023}. In this situation, it is unclear whether a reported ranking reflects the aggregation strategies or the protocol used to evaluate them. It is also unclear whether conclusions drawn from federations simulated by splitting one hospital into care units, a common shortcut, carry over to federations of independent hospitals.

This study addresses these questions with a controlled comparison. It asks, first, which aggregation strategy performs best under a protocol that treats every strategy alike, and whether the answer depends on the kind of federation (RQ1); second, how close federated training comes to centralized training once the protocol is controlled (RQ2); and third, how much common protocol choices change the conclusions (RQ3). To answer them, seven strategies (FedAvg, FedProx, SCAFFOLD, FedAdagrad, FedAdam, FedYogi and FedCluster) were trained with the same pipeline and predictors on two federations: 12 care units of one hospital in MIMIC-IV and 49 hospitals in eICU-CRD. Each strategy's hyperparameters were tuned on validation data; every model was selected by the same validation rule; every comparison was repeated over 10 seeds and tested with procedures designed for several methods; and calibration was assessed alongside discrimination.

The study makes four contributions. It provides a controlled comparison of seven aggregation strategies on an intra-hospital and an inter-hospital federation built with a single pipeline, a design that separates the effect of the federation structure from that of data preparation. It quantifies the effect of four protocol choices, namely discharge-coded predictors, client-level oversampling, the round at which the model is evaluated, and the tuning of hyperparameters, and shows that three of them change discrimination by more than the choice of strategy does in the care-unit federation, and one of them in both federations. It identifies SCAFFOLD, absent from previous clinical comparisons, as the most accurate strategy in both federations, together with the condition under which it works, plain local gradient descent with a tuned learning rate, and uses a control strategy to show that its advantage comes mainly from its drift correction rather than from that change of optimizer. Finally, it shows that federated training matches centralized training on the multi-hospital federation and that a simple recalibration step repairs the risk estimates distorted by imbalance corrections. The remainder of the paper reviews related work (Section~\ref{sec:related}), describes the methods (Section~\ref{sec:methods}), presents the results (Section~\ref{sec:results}), discusses them (Section~\ref{sec:discussion}), and concludes (Section~\ref{sec:conclusion}).

\section{Related Work}
\label{sec:related}
\subsection{Federated learning for EHR-based prediction}
Federated learning was proposed for training models on data that cannot leave the devices or institutions that hold it~\cite{mcmahan2017}, and health care soon became one of its main applications~\cite{rieke2020}. The first studies on electronic health records showed that useful models could be trained without pooling the records: Brisimi et al.~\cite{brisimi2018} distributed the records of one medical center among five or ten simulated sites to predict cardiac hospitalizations with a decentralized sparse support vector machine, and Huang et al.~\cite{huang2019cbfl} trained one model per cluster of similar patients across 50 eICU-CRD hospitals, which predicted mortality and length of stay better than plain federated averaging. Since then, the eICU-CRD database, which groups ICU stays by hospital, has served as the usual testbed, because its hospitals form a natural federation. Pfohl et al.~\cite{pfohl2019} trained models for in-hospital mortality and prolonged length of stay across 31 eICU-CRD hospitals with FedAvg, with and without differential privacy, and found that federated training approached centralized training but that strong privacy guarantees were harder to obtain in the federated setting. When multi-hospital data are not available, federations are simulated by partitioning a single-center database: Lee and Shin~\cite{lee2020fl} split MIMIC-III at random into three clients to predict in-hospital mortality, and Mondrejevski et al.~\cite{mondrejevski2022} trained recurrent and convolutional models for ICU mortality on two to eight random MIMIC-III clients; both found federated performance close to centralized performance. Vaid et al.~\cite{vaid2021}, in contrast, used the records of five hospitals of one health system to predict seven-day mortality in COVID-19 and found that federated models outperformed models trained at each hospital alone, and Kerkouche et al.~\cite{kerkouche2021}, with 314 hospitals of the Premier Healthcare Database, showed that differentially private and compressed updates cost only a moderate loss of discrimination in in-hospital mortality prediction. Rajendran et al.~\cite{rajendran2023} examined how heterogeneity between seven hospitals affects federated models for acute kidney injury and sepsis, again with FedAvg. Dang et al.~\cite{dang2022} compared six aggregation strategies on 20 hospitals for mortality and acute kidney injury and concluded that simple averaging performed as well as more elaborate strategies, and Zhu et al.~\cite{zhu2025fedweight} proposed a re-weighting method for covariate shift between hospitals that they compared with FedAvg and FedProx. Benchmark suites such as FLamby~\cite{duterrail2022} offer naturally partitioned medical datasets, but none include EHR-based mortality prediction.

Table~\ref{tab:related} summarizes how the peer-reviewed studies among them evaluated the strategies. Most compared only FedAvg, or FedAvg and a proposed variant, with local and centralized training, under fixed hyperparameters and a single split or cross-validation; one selected the reported round by its performance on the test data, and the one that compared several aggregation strategies did not tune them, used different selection rules for federated and centralized models, and reported a single run. None reported the calibration of the predicted risks. When multi-hospital data are unavailable, federations are often simulated by splitting a single hospital's records at random or by care unit; whether conclusions obtained on such intra-hospital federations hold for federations of independent hospitals has not been examined.

\begin{table*}[t]\centering
\caption{Peer-reviewed studies of federated learning for EHR-based prediction compared with this study. AKI: acute kidney injury; LOS: length of stay; CBFL: community-based federated learning; cPDS: cluster primal--dual splitting; DP: differential privacy; SVM: support vector machine; CI: confidence interval.}
\label{tab:related}
\footnotesize
\setlength{\tabcolsep}{3pt}
\adjustbox{max width=\linewidth}{
\begin{tabular}{@{}lllllll@{}}
\toprule
Study & Data and clients & Outcomes & Strategies compared & Hyperparameters & Model selection & Repetitions and tests \\
\midrule
Brisimi et al.~\cite{brisimi2018} & one center, 5 or 10 & cardiac & cPDS vs other & not reported & until convergence & one random split \\
 & simulated sites & hospitalization & decentralized SVM solvers & & & \\
Huang et al.~\cite{huang2019cbfl} & eICU-CRD, 50 hospitals & mortality, ICU LOS & FedAvg, CBFL & fixed (defaults) & until convergence & one split \\
Lee and Shin~\cite{lee2020fl} & MIMIC-III, 3 random & in-hospital mortality & FedAvg & fixed & last round & one split; \\
 & clients & & & & & resampled CIs \\
Vaid et al.~\cite{vaid2021} & 5 hospitals of one & 7-day mortality & FedAvg (with and & tuned, procedure & not reported & 490 random splits; \\
 & health system & (COVID-19) & without noise) & not described & & CIs \\
Kerkouche et al.~\cite{kerkouche2021} & Premier database, & in-hospital mortality & FedAvg, sign-compressed & grid search & best test-set round & 5 random splits; \\
 & 314 hospitals & & FL (with and without DP) & & & min--max \\
Dang et al.~\cite{dang2022} & eICU-CRD, 20 hospitals & mortality, AKI & FedAvg, FedAvgM, FedProx, & not reported & FL: last round; & one run; DeLong and \\
 & & & FedAdagrad, FedAdam, FedYogi & & centralized: best epoch & logit CIs \\
Mondrejevski et al.~\cite{mondrejevski2022} & MIMIC-III, 2, 4 or 8 & ICU mortality & FedAvg & fixed & early stopping on & five-fold cross-validation \\
 & random clients & & & & validation loss & \\
Rajendran et al.~\cite{rajendran2023} & eICU-CRD, 7 hospitals & AKI, sepsis & FedAvg & fixed & fixed number of epochs & five-fold cross-validation \\
Zhu et al.~\cite{zhu2025fedweight} & eICU-CRD hospitals; & ICU mortality and & FedAvg, FedProx, FedWeight & grid search for FedProx & early stopping on & several seeds; rank-sum \\
 & eICU-CRD to MIMIC-III & three other outcomes & & & target validation & test vs FedAvg \\
\midrule
This study & MIMIC-IV, 12 care units; & in-hospital mortality & FedAvg, FedProx, SCAFFOLD, & grid search per strategy & best validation round, & 10 seeds; Friedman \\
 & eICU-CRD, 49 hospitals & & FedAdagrad, FedAdam, FedYogi, FedCluster & on validation & same rule for all & and Holm-corrected tests \\
\bottomrule
\end{tabular}}\\[2pt]
\parbox{\linewidth}{\scriptsize None of the previous studies reported the calibration of the predicted risks; this study reports Brier score, calibration intercept and slope and expected calibration error, before and after recalibration.}
\end{table*}
\subsection{Aggregation strategies}
\label{sec:rw-strategies}
The strategies compared in this study were chosen to represent the four main ways federated optimization addresses heterogeneous clients. FedAvg~\cite{mcmahan2017} averages local models and is the reference method. FedProx~\cite{li2020fedprox} and SCAFFOLD~\cite{karimireddy2020} correct client drift, the first by penalizing the distance between local and global models and the second by subtracting an estimate of each client's drift from its gradients. FedAdagrad, FedAdam, and FedYogi~\cite{reddi2021} keep plain local training but replace the server's averaging step with an adaptive optimizer. FedCluster, finally, stands for robust aggregation, in a variant implemented for this study (Section~\ref{sec:methods}): it clusters the client updates by direction and averages only the largest cluster, discarding updates that point elsewhere. Other drift-correction methods, such as FedNova~\cite{wang2020fednova}, which normalizes for unequal numbers of local steps, and FedDyn~\cite{acar2021feddyn}, which adds a dynamic regularizer, address the same problem as FedProx and SCAFFOLD and were left out to keep the comparison focused; of the clinical studies in Table~\ref{tab:related}, none evaluated SCAFFOLD.
\subsection{Evaluation pitfalls in clinical machine learning}
Several practices that affect comparisons of prediction models are well-documented outside federated learning. Leakage, the use of information that would not be available when the prediction is made, is a common source of overoptimistic results in machine learning~\cite{kapoor2023}. Correcting class imbalance by resampling or weighting does not improve discrimination but systematically overestimates the risk of the minority class~\cite{goorbergh2022}, which matters because clinical use depends on calibrated probabilities~\cite{vancalster2019}. In federated learning, hyperparameter tuning is itself difficult and noisy~\cite{kuo2023}, so that untuned comparisons may favor whichever method the default settings happen to suit. Finally, comparing several methods over repeated runs requires tests that control the error rate of multiple comparisons~\cite{demsar2006}, and reporting guidelines for prediction models now ask for calibration, uncertainty and the handling of imbalance to be reported explicitly~\cite{collins2024}. The protocol described in the next section was designed to address each of these points and measure how much each matters.

\section{Materials and Methods}
\label{sec:methods}
\subsection{Data sources and cohorts}
\begin{figure*}[t]\centering
\begin{subfigure}{\textwidth}\centering
\adjustbox{max width=\textwidth}{\begin{tikzpicture}[
    font=\sffamily\scriptsize,
    node distance=3mm and 4mm,
    box/.style={draw=black!45, line width=0.4pt, rounded corners=1.5pt, fill=white,
                align=center, inner sep=3pt},
    fam/.style={box, draw=rita, line width=0.7pt, fill=rita!6, text width=31mm, minimum height=11mm},
    client/.style={box, text width=33mm, minimum height=24mm},
    db/.style={cylinder, shape border rotate=90, aspect=0.25, draw=black!45, fill=black!4,
               minimum width=7mm, minimum height=7mm, inner sep=1pt, align=center},
    arr/.style={-{Stealth[length=1.8mm]}, line width=0.6pt, draw=black!60},
  ]
  \definecolor{rita}{HTML}{2a78d6}

  \node[fam] (f1) {\textbf{Weighted averaging}\\FedAvg, FedProx};
  \node[fam, right=of f1] (f2) {\textbf{Drift-corrected step}\\SCAFFOLD ($\eta_g$, control variate $c$)};
  \node[fam, right=of f2] (f3) {\textbf{Adaptive server step}\\FedAdagrad, FedAdam, FedYogi ($\eta$, $\tau$)};
  \node[fam, right=of f3] (f4) {\textbf{Robust aggregation}\\FedCluster\\(\mbox{DBSCAN} quantile $q$)};
  \node[draw=black!60, line width=0.6pt, rounded corners=2pt, inner sep=4pt,
        fit=(f1)(f2)(f3)(f4), label={[font=\sffamily\scriptsize\bfseries, anchor=south west]north west:Server: one aggregation rule per run}] (server) {};

  \node[client, below=17mm of f1.south, anchor=north] (c1) {};
  \node[client, below=17mm of f2.south, anchor=north] (c2) {};
  \node[box, below=17mm of f3.south, anchor=north, text width=33mm, minimum height=24mm, draw=none] (cdots) {\Large$\cdots$};
  \node[client, below=17mm of f4.south, anchor=north] (ck) {};
  \foreach \c/\name in {c1/Client 1, c2/Client 2, ck/Client $K$} {
    \node[anchor=north, font=\sffamily\scriptsize\bfseries] at (\c.north) {\name};
    \node[db, anchor=west] (\c db) at ([xshift=1.3mm, yshift=-2mm]\c.west) {\tiny local\\\tiny data};
    \node[anchor=west, align=left, font=\sffamily\scriptsize] at ([xshift=0.8mm]\c db.east)
      {MLP 128--64\\3 local epochs\\weighted loss\\FedProx: + prox.\ term\\SCAFFOLD: + $c-c_k$};
  }

  \foreach \c in {c1, c2, ck} {
    \draw[arr] ([xshift=-5mm]server.south -| \c.north) -- node[left, font=\sffamily\tiny, align=right] {$w^t$} ([xshift=-5mm]\c.north);
    \draw[arr] ([xshift=5mm]\c.north) -- node[right, font=\sffamily\tiny, align=left] {$w_k^{t+1}$} ([xshift=5mm]server.south -| \c.north);
  }

  \draw[decorate, decoration={brace, mirror, amplitude=4pt}, draw=black!50]
    ([yshift=-1.5mm]c1.south west) -- ([yshift=-1.5mm]ck.south east)
    node[midway, below=4pt, align=center, font=\sffamily\scriptsize]
    {records never leave the client \quad$\cdot$\quad MIMIC-IV: $K=12$ care units of one hospital \quad$\cdot$\quad eICU-CRD: $K=49$ hospitals};

  \node[align=center, font=\sffamily\scriptsize, text=black!65] at ($(server.south)!0.5!(cdots.north) + (f3.center |- 0,0) - (server.center |- 0,0)$)
    {each round $t=1,\dots,50$:\\$w^t$ down, $w_k^{t+1}$ up;\\the round with the best\\validation AUC-PR is kept};
\end{tikzpicture}}
\caption{Federated architecture. In each round, the server sends the global model $w^t$ to the $K$ clients; each client trains it on its own records, and the server combines the resulting models $w_k^{t+1}$ using the aggregation rule of the strategy under study.}
\label{fig:architecture}
\end{subfigure}\\[4mm]
\begin{subfigure}{\textwidth}\centering
\adjustbox{max width=\textwidth}{\begin{tikzpicture}[
    font=\sffamily\scriptsize,
    node distance=4mm and 5mm,
    box/.style={draw=black!45, line width=0.4pt, rounded corners=1.5pt, fill=white,
                align=center, inner sep=3pt, minimum height=13mm, text width=32mm},
    key/.style={box, draw=rita, line width=0.8pt, fill=rita!6},
    arr/.style={-{Stealth[length=1.8mm]}, line width=0.5pt, draw=black!60},
  ]
  \definecolor{rita}{HTML}{2a78d6}

  \node[box] (data) {\textbf{Data}\\MIMIC-IV v3.1\\eICU-CRD v2.0};
  \node[box, right=of data] (cohort) {\textbf{Cohort}\\one row per hospital stay\\features from the first 24\,h};
  \node[key, right=of cohort] (clients) {\textbf{Clients}\\sites with $\geq$100 deaths\\12 care units $\mid$ 49 hospitals};
  \node[box, right=of clients] (split) {\textbf{Patient-level split}\\train / val. / test = 70 / 10 / 20\%\\repeated for 10 seeds};

  \node[key, below=6mm of split] (train) {\textbf{Federated training}\\7 aggregation strategies\\50 rounds $\times$ 3 local epochs\\class-weighted loss};
  \node[key, left=of train] (select) {\textbf{Model selection}\\global model at the round\\with best validation AUC-PR};
  \node[key, left=of select] (recal) {\textbf{Recalibration}\\Platt scaling\\fitted on validation};
  \node[box, left=of recal] (eval) {\textbf{Test evaluation}\\AUC-ROC, AUC-PR, Brier\\Friedman + Holm tests};

  \node[box, below=of train, dashed, minimum height=9mm] (tune) {\textbf{Per-strategy tuning}\\grid search on validation};
  \node[box, below=of eval, minimum height=9mm] (central) {\textbf{Centralized baselines}\\same split, pooled training};

  \draw[arr] (data) -- (cohort);
  \draw[arr] (cohort) -- (clients);
  \draw[arr] (clients) -- (split);
  \draw[arr] (split) -- (train);
  \draw[arr] (train) -- (select);
  \draw[arr] (select) -- (recal);
  \draw[arr] (recal) -- (eval);
  \draw[arr, dashed] (tune) -- (train);
  \draw[arr] (central) -- (eval);
\end{tikzpicture}}
\caption{Evaluation protocol. Blue boxes denote the protocol choices whose effects the study measures; the same split and test set are used for both the federated strategies and the centralized baselines.}
\label{fig:protocol}
\end{subfigure}
\caption{System and protocol of the study.}
\label{fig:system}
\end{figure*}
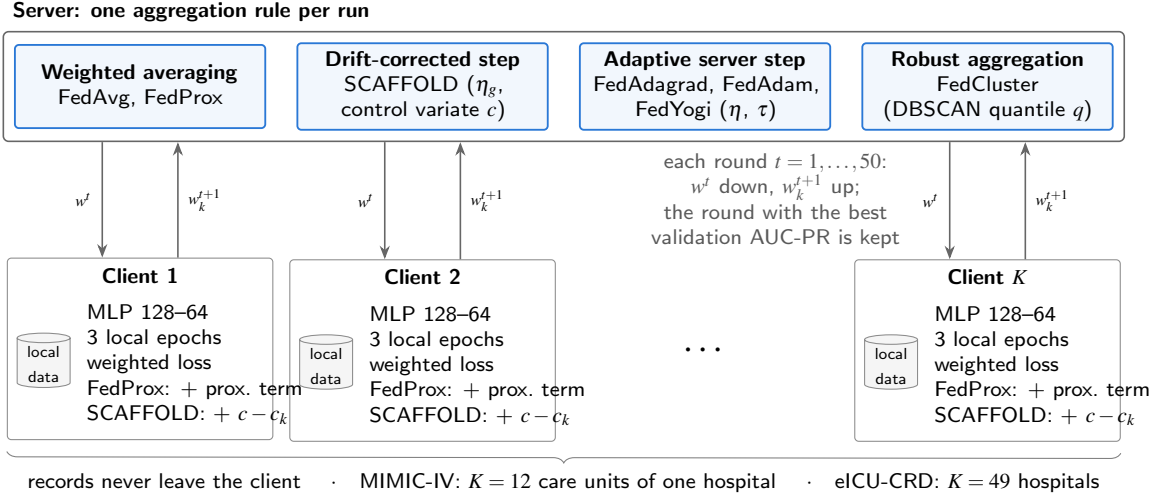
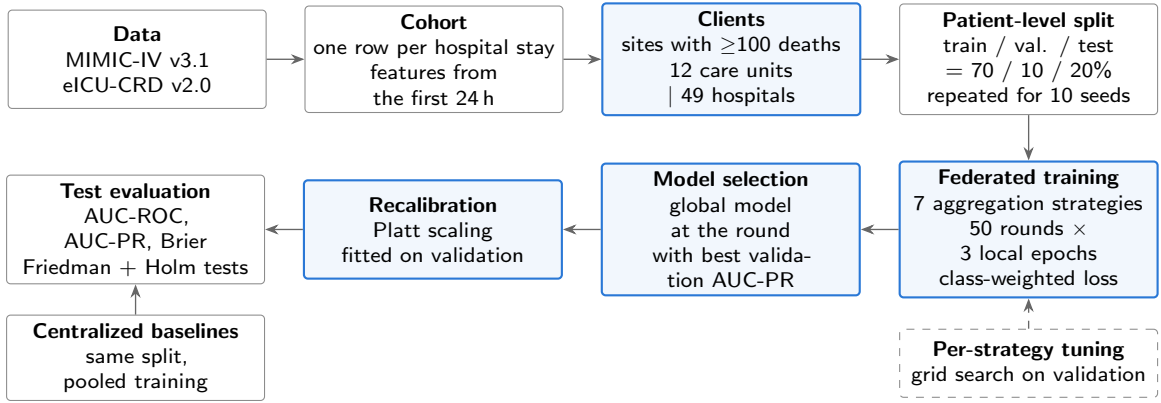

This section describes the study in the order of the evaluation protocol summarized in Figure~\ref{fig:protocol}, which operates on the federated architecture of Figure~\ref{fig:architecture}: the data and cohorts, the construction of the two federations, the splitting and preprocessing of the records, the model and the aggregation strategies, the training and selection procedures, the centralized baselines and, finally, the evaluation and statistical analysis. The report follows the items of the TRIPOD+AI statement that apply to a methodological comparison of prediction models~\cite{collins2024}, together with its TRIPOD-Cluster extension for models developed on data from several sites~\cite{debray2023}.

The study used two publicly available, de-identified databases distributed by PhysioNet~\cite{goldberger2000}. The first is MIMIC-IV, version~3.1~\cite{johnson2023,mimiciv31}, which contains the electronic health records of patients admitted to the Beth Israel Deaconess Medical Center in Boston, United States, between 2008 and 2022. The second is the eICU Collaborative Research Database (eICU-CRD), version~2.0~\cite{pollard2018,eicu20}, which contains intensive care unit (ICU) stays from 208 hospitals across the United States in 2014 and 2015. Access to both databases was obtained through PhysioNet's credentialing process and is governed by their data use agreements; because the data are de-identified and the analysis was retrospective, no additional ethical approval was required.

In both databases, the unit of analysis was the hospital stay of an adult patient (18 years or older), and the outcome was in-hospital death. In MIMIC-IV, every adult admission was included, yielding 546{,}028 hospital stays from about 223{,}000 patients, and each stay was assigned to the care unit where it began. In eICU-CRD, each hospital stay was described by its first ICU stay; stays without a recorded discharge status were excluded, and ages recorded as ``$>89$'' were set to 90, resulting in 164{,}322 hospital stays from 137{,}773 patients across 207 hospitals (the 208th hospital contributed 13 stays, none with a recorded discharge status).

The predictors were restricted to information available within 24 hours, because the models are intended to estimate risk early in the stay. They comprised age, sex and race or ethnicity; insurance type in MIMIC-IV, and ICU type and hospital admission source in eICU-CRD; and a panel of 11 laboratory measurements (creatinine, blood urea nitrogen, sodium, potassium, glucose, white blood cell count, hemoglobin, hematocrit, platelet count, bicarbonate and lactate), each averaged over the first 24 hours after hospital admission in MIMIC-IV and after ICU admission in eICU-CRD. The panel is deliberately the same in both databases: most MIMIC-IV admissions never reach an ICU and therefore have no ICU vital signs, so a shared feature set is what allows the two federations to be compared on equal terms. Counts of the diagnosis and procedure codes of the stay, a convenient summary of comorbidity, were excluded because those codes are assigned at discharge; their effect is quantified in Section~\ref{sec:protocol}. To examine whether the conclusions depend on this restricted feature set, a sensitivity analysis on eICU-CRD added 11 first-day physiological variables from the APACHE~IV dataset (worst heart rate, mean arterial pressure, temperature and respiratory rate, the three Glasgow Coma Scale components, urine output, and indicators of intubation, mechanical ventilation and dialysis); urine outputs outside 0--20{,}000\,mL per day were treated as missing. Table~\ref{tab:cohorts} summarizes both cohorts.
\begin{table}[t]\centering
\caption{Cohorts and federations. Client figures refer to the sites that meet the eligibility rule.}
\label{tab:cohorts}
\footnotesize
\adjustbox{max width=\linewidth}{\begin{tabular}{@{}lrr@{}}
\toprule
 & MIMIC-IV & eICU-CRD \\
\midrule
Hospital stays (all sites) & 546,028 & 164,322 \\
Hospital stays (clients) & 475,570 & 99,987 \\
Patients (clients) & 206,420 & 85,917 \\
In-hospital mortality & 2.35\% & 9.97\% \\
Age, mean (SD), years & 58.2 (18.5) & 63.0 (17.0) \\
Female & 50.1\% & 45.6\% \\
Clients (sites with $\geq$100 deaths) & 12 & 49 \\
Stays per client, min--max & 875--374,252 & 808--5,499 \\
Largest client's share of stays & 78.7\% & 5.5\% \\
Mortality per client, min--max & 0.3--20.3\% & 5.3--17.5\% \\
Lab panel coverage, min--max\textsuperscript{a} & 13--68\% & 26--89\% \\
\bottomrule
\end{tabular}
}\\[2pt]\parbox{\linewidth}{\scriptsize \textsuperscript{a}Share of client stays with each of the 11 laboratory measurements in the first 24\,h; the lowest is lactate in both databases.}
\end{table}
\subsection{Federations}
The two databases support two different kinds of federation, and the study uses both on purpose. In both cases, a site, that is, a care unit in MIMIC-IV or a hospital in eICU-CRD, became a client when it had at least 100 in-hospital deaths. A rule based on events was preferred to one based on size, which would be arbitrary here: the fifth- and sixth-largest care units in MIMIC-IV differ by only about 1{,}000 stays, and the five largest include a unit with a single death in 20{,}337 stays. With at least 100 deaths, even the smallest client keeps, under the split described below, about 65 deaths for training and 25 for testing, enough to learn from its positive cases and to be evaluated on its own.

\begin{figure*}[t]\centering
\includegraphics[width=\textwidth]{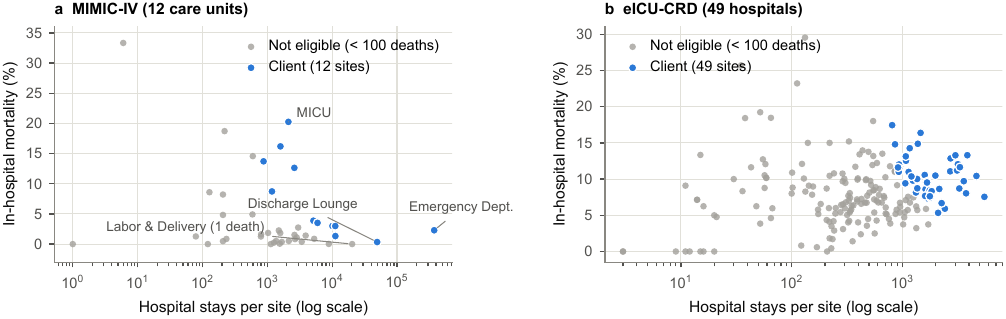}
\caption{Every site of the two databases by number of hospital stays and in-hospital mortality. Blue: sites that meet the eligibility rule ($\geq$100 deaths) and become clients. (a) MIMIC-IV care units: one dominant client (the Emergency Department) and mortality from 0.3\% to 20\%. (b) eICU-CRD hospitals.}
\label{fig:federations}
\end{figure*}

Applied to MIMIC-IV, the rule selected 12 care units that hold 87\% of the hospital stays (Figure~\ref{fig:federations}a). This is an intra-hospital federation with an extreme imbalance: the Emergency Department alone accounts for 78.7\% of clients' stays, and mortality ranges from 0.3\% to 20.3\% across units. Applied to eICU-CRD, the rule selected 49 hospitals (Figure~\ref{fig:federations}b), an inter-hospital federation of institutions with their own case mix and practices, in which the largest client holds only 5.5\% of the stays and mortality ranges from 5.3\% to 17.5\%. Four variants examine whether the results depend on these choices: one extra client that pools all remaining sites, in both databases; the Emergency Department split at random into four clients by patient, in MIMIC-IV; the five largest care units, in MIMIC-IV; and a stricter threshold of 200 deaths, which leaves 17 hospitals in eICU-CRD.
\subsection{Splitting, preprocessing, and class imbalance}
Once the clients were defined, the records were split so that no information could cross from training to evaluation. The split was made at the patient level: each patient, with all of his or her stays, was assigned to the training (70\%), validation (10\%), or test (20\%) partition, stratified by whether the patient died in any stay, and each client's partitions were the corresponding subsets of its stays. The union of the clients' validation partitions formed the global validation set, and the union of their test partitions formed the global test set, which every federated strategy and every centralized baseline shared within a seed. The whole procedure was repeated with ten seeds (42 to 51), each producing a different split.

Preprocessing followed the same rule. Missing laboratory values were imputed with the median of the pooled training rows and flagged with a binary indicator per measurement, because a test that was not ordered is itself informative; categorical variables were one-hot encoded; and numerical variables were standardized using statistics computed only on the training rows. These statistics are the only quantities shared across clients beyond model updates; in a deployment, they would be obtained by a secure aggregation of client summaries.

In-hospital death is a rare outcome (2.35\% among the MIMIC-IV clients and 9.97\% among the eICU-CRD clients), and the treatment of this imbalance is one of the protocol choices the study examines. In the main setting, each client weighted the loss of its positive cases by the ratio of negative to positive training cases, capped at 20, so that a client with very few deaths could not dominate an aggregation step. Two ablations replaced this treatment: SMOTE-Tomek~\cite{chawla2002,batista2004} applied only to each client's training partition, followed by the same weighting computed on the resampled rows; and no correction at all.
\subsection{Model and aggregation strategies}
All federated strategies trained the same model, a multilayer perceptron with two hidden layers of 128 and 64 rectified linear units and a sigmoid output, chosen because tabular electronic health record tasks of this size rarely benefit from deeper networks; a logistic regression was also trained as an architecture check. The seven strategies cover the four families described in Section~\ref{sec:rw-strategies}. In all of them, client $k$ holds $n_k$ training rows, $n=\sum_k n_k$, and in round $t$ each participating client starts from the global model $w^t$, trains locally, and returns $w_k^{t+1}$ (Figure~\ref{fig:architecture}); the strategies differ in the local objective, in the server's update rule, or in both.

FedAvg~\cite{mcmahan2017}, the reference method, replaces the global model by the size-weighted average of the local ones,
\begin{equation}
  w^{t+1}=\sum_{k} \frac{n_k}{n}\,w_k^{t+1}.
  \label{eq:fedavg}
\end{equation}
Two strategies correct the drift of local models away from the global one. FedProx~\cite{li2020fedprox} adds the proximal term $(\mu/2)\lVert w-w^t\rVert^2$ to each local objective and aggregates with Equation~\eqref{eq:fedavg}. SCAFFOLD~\cite{karimireddy2020} adds the correction $c-c_k$ to every local gradient, where the control variate $c_k$ estimates the gradient of client $k$ and $c=\sum_k (n_k/n)\,c_k$ estimates the global one. In this study, $c_k$ was recomputed in each round as the full-batch gradient of the client's loss at $w^t$ (option~I of the original method), and the server moved the global model by a step $\eta_g$ towards the average of Equation~\eqref{eq:fedavg},
\begin{equation}
  w^{t+1}=w^t+\eta_g\Big(\sum_{k} \frac{n_k}{n}\,w_k^{t+1}-w^t\Big).
\end{equation}
SCAFFOLD trains locally with stochastic gradient descent (SGD) at its own learning rate $\eta_l$, as in its original formulation; with the adaptive Adam optimizer~\cite{kingma2015} used by the other strategies, the per-coordinate normalization turns the correction, which is constant within a round, into a fixed drift at every step, and on the MIMIC-IV federation, where the Emergency Department performs more than 1{,}500 local steps per round, this made the model collapse within three rounds.

Three strategies apply adaptive optimization at the server~\cite{reddi2021}. They treat the average client update $\Delta_t=\sum_k (n_k/n)\,w_k^{t+1}-w^t$ as a pseudo-gradient and take an adaptive step,
\begin{equation}
  w^{t+1}=w^t+\eta\,\frac{m_t}{\sqrt{v_t}+\tau},
  \label{eq:adaptive}
\end{equation}
where $\eta$ is the server learning rate and $\tau$ a stability constant. FedAdagrad uses $m_t=\Delta_t$ and accumulates $v_t=v_{t-1}+\Delta_t^2$. FedAdam and FedYogi use the momentum $m_t=\beta_1 m_{t-1}+(1-\beta_1)\Delta_t$ and differ in the second moment,
\begin{align}
  v_t &= \beta_2 v_{t-1}+(1-\beta_2)\,\Delta_t^2 && \text{(FedAdam)},\\
  v_t &= v_{t-1}-(1-\beta_2)\,\Delta_t^2\,\operatorname{sign}\!\big(v_{t-1}-\Delta_t^2\big) && \text{(FedYogi)},
\end{align}
with $\beta_1=0.9$ and $\beta_2=0.99$. The last family discards anomalous updates. FedCluster, the name used here for a clustering-based robust aggregation implemented for this study (unrelated to the cluster-cycling method of the same name~\cite{chen2020fedcluster}), normalizes the client updates $w_k^{t+1}-w^t$ to unit length, clusters them with DBSCAN~\cite{ester1996} using a radius equal to a quantile $q$ of their pairwise distances, and moves the global model by the unweighted mean update of the largest cluster, or of all clients when no cluster forms. Personalized federated learning was outside the scope of the comparison and is examined only as local fine-tuning of the selected global model (one epoch on each client's training data).
\subsection{Training, tuning and model selection}
Training followed a fixed schedule, so the strategies differed only in how they aggregated. Each run lasted 50 communication rounds, in which, unless stated otherwise, all clients participated. In each round, a client trained for three local epochs with mini-batches of 512 rows, using Adam with a learning rate of $10^{-3}$ (SGD with a tuned learning rate $\eta_l$ for SCAFFOLD and for the FedAvg-SGD control). Sensitivity analyses used one or five local epochs, 100 rounds, and partial participation, in which 50\% or 20\% of the clients were sampled uniformly at random in each round, with the same sample sequence for every strategy within a seed.

The hyperparameters specific to each strategy were tuned rather than fixed, because a shared setting favors whichever method it happens to suit. For each dataset, a grid search on the main configuration with seed 42 selected the value that maximized validation AUC-PR, with ties broken by validation AUC-ROC (Table~\ref{tab:tuning} in the appendix). The grids covered $\mu$ for FedProx, the server learning rate $\eta$ and constant $\tau$ for the adaptive strategies, the local learning rate and $\eta_g$ for SCAFFOLD, and the quantile $q$ for FedCluster; when the optimum fell on the edge of a grid, the grid was extended. The grids therefore differed in size, from a single configuration for FedAvg, which has no strategy-specific hyperparameter, to 12 configurations on MIMIC-IV and 14 on eICU-CRD for SCAFFOLD, whose local learning rate was tuned as well. Because SCAFFOLD was the only strategy whose local optimizer was tuned, a control strategy, FedAvg-SGD, combined the aggregation of FedAvg with local SGD tuned on the same learning-rate grid; the difference between SCAFFOLD and FedAvg-SGD isolates the effect of the control variates from that of the local optimizer, and this pre-specified contrast was tested on its own, outside the families corrected by Holm's method. To measure what tuning contributes, the tuned values were also compared with a common default setting ($\mu=0.1$; $\eta=0.01$ and $\tau=10^{-3}$ for the adaptive strategies; $\eta_l=0.03$ and $\eta_g=1$ for SCAFFOLD; $q=0.5$ for FedCluster). The selected values were then used, unchanged, in every experiment on that dataset, including the eICU-CRD sensitivity analysis with physiological variables. Because seed 42 was used for tuning, the main comparisons were also repeated without it.

The model reported for each run was also selected on the validation set. After every round the server computed the validation AUC-PR of the global model and kept the model of the best round, so that the number of rounds acts as an upper bound rather than as a choice that favors the strategies peaking near it; the model of the last round was also stored for comparison. Because both class weighting and resampling distort predicted risks~\cite{goorbergh2022}, the selected model was finally recalibrated using Platt scaling~\cite{platt1999}, a logistic regression of the outcome on the logit of the predicted probability, fitted on the validation set; isotonic regression~\cite{zadrozny2002} was evaluated as an alternative. Both steps use only validation data; in this study they were computed on the pooled validation set, and Section~\ref{sec:discussion} discusses how they would be carried out in a deployment.
\subsection{Centralized baselines}
To measure what federation costs, the same multilayer perceptron was also trained on the pooled training rows of all clients, as if the data could be centralized. Three variants matched the imbalance treatments of the federated strategies: no correction, class weighting with the same cap, and SMOTE-Tomek applied to the pooled rows. Each centralized model was trained with Adam (learning rate $10^{-3}$, weight decay $10^{-5}$, batches of 512) for up to 100 epochs, with early stopping when the validation loss had not improved for five epochs, and was evaluated on the same test set as the federated strategies of the same seed.
\subsection{Evaluation and statistical analysis}
Discrimination on the global test set was the primary outcome, measured by the area under the receiver operating characteristic curve (AUC-ROC) and by the area under the precision--recall curve (AUC-PR, computed as average precision), which is the more informative of the two when positives are rare. Calibration was assessed by the Brier score, the calibration intercept (calibration-in-the-large, ideally 0), the calibration slope (ideally 1), and the expected calibration error over ten equal-frequency bins~\cite{vancalster2019}. Each client was also evaluated on its own test partition, and performance was compared across sociodemographic subgroups (sex, age under or over 65 years, and race or ethnicity) to examine fairness. Threshold-dependent measures, used only to examine how the predicted probabilities translate into decisions, were computed at the threshold that maximized the F1 score on the validation set.

The statistical analysis treated the seed as the unit of replication, since within a seed, all strategies share the split and the test set. Following the procedure recommended for comparing several methods over repeated runs~\cite{demsar2006}, a Friedman test~\cite{friedman1937} first compared the seven federated strategies; the strategy with the best mean rank was then compared with each of the others by the paired $t$-test and the Wilcoxon signed-rank test~\cite{wilcoxon1945}, with Holm's correction~\cite{holm1979} over the six comparisons of each metric, and with each centralized baseline as a separate family. Effects are reported as mean paired differences with 95\% confidence intervals. With ten seeds, the smallest attainable two-sided Wilcoxon $p$-value is $2/2^{10}\approx0.002$, so correcting over all pairs of strategies at once would make significance unreachable; with five seeds, the floor is higher still, and for those experiments, the paired $t$-test and the confidence intervals are reported. Table~\ref{tab:experiments} lists the experiments.

The study was implemented in Python~3.12 with PyTorch~2.13, scikit-learn~1.9 and imbalanced-learn~0.14, as a synchronous simulation of federated training on a single machine; the code is publicly available (see Data and code availability). The results of this protocol are presented next, starting with the comparison of the strategies.
\begin{table}[t]\centering
\caption{Experiments. All federated runs use the per-strategy hyperparameters selected on the main configuration of the same dataset.}
\label{tab:experiments}
\footnotesize
\adjustbox{max width=\linewidth}{
\begin{tabular}{@{}llc@{}}
\toprule
Experiment & What changes & Seeds \\
\midrule
Main & class-weighted loss, sites with $\geq$100 deaths & 10 \\
Imbalance ablation & SMOTE-Tomek per client; no correction & 10 \\
Federation variants & + pooled remaining sites; ED split in 4 (MIMIC-IV); & 5 \\
 & five largest units (MIMIC-IV); $\geq$200 deaths (eICU-CRD) & \\
Local epochs & 1 or 5 instead of 3 & 3 \\
Participation & 50\% (both), 20\% (eICU-CRD) of clients per round & 3 \\
Rounds & 100 instead of 50 (MIMIC-IV) & 1 \\
Architecture & logistic regression instead of the MLP & 3 \\
Richer features & eICU-CRD + 11 day-1 APACHE variables & 5 \\
Control for SCAFFOLD & FedAvg with local SGD, learning rate tuned on SCAFFOLD's grid & 10 \\
Personalization & 1 epoch of local fine-tuning of the main models & 10 \\
Recalibration & Platt and isotonic, fitted on validation & 10 \\
\bottomrule
\end{tabular}}
\end{table}

\section{Results}
\label{sec:results}
\subsection{Aggregation strategies in the two federations (RQ1)}
\begin{figure*}[t]\centering
\includegraphics[width=\textwidth]{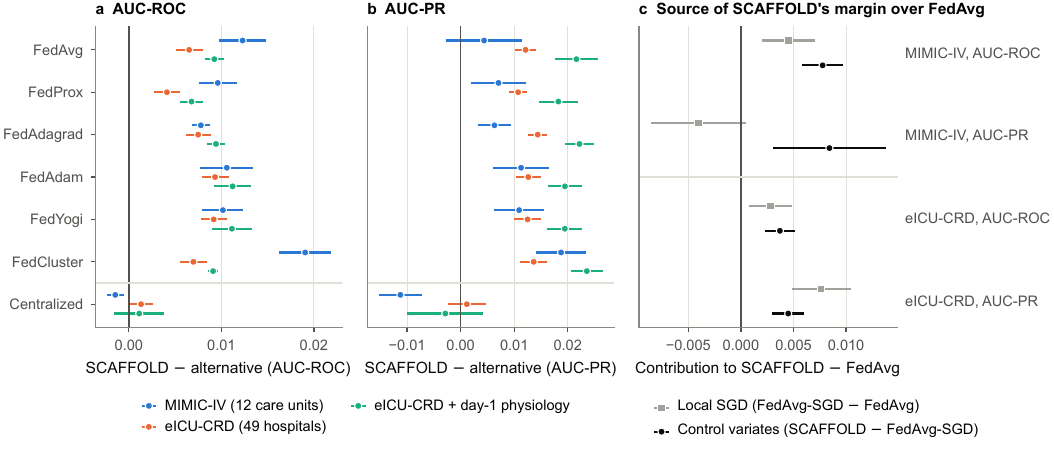}
\caption{Paired differences by seed (mean and 95\% confidence interval). (a, b) SCAFFOLD minus each alternative (10 seeds; 5 for eICU-CRD with day-1 physiology); values to the right of zero favor SCAFFOLD, and ``Centralized'' is the centralized model without imbalance correction. (c) SCAFFOLD's margin over FedAvg split into the contribution of local SGD with a tuned learning rate (FedAvg-SGD minus FedAvg) and that of the control variates (SCAFFOLD minus FedAvg-SGD); the two contributions add up to the FedAvg row of panels a and b.}
\label{fig:paired}
\end{figure*}

The results are presented in the order of the research questions, starting with the comparison of the seven strategies under the protocol of Section~\ref{sec:methods} (Table~\ref{tab:main} and Figure~\ref{fig:paired}). In the MIMIC-IV care-unit federation, SCAFFOLD reached the highest AUC-ROC, $0.886 \pm 0.002$, and ranked first in all ten seeds (Friedman $\chi^2=44.2$, $p<0.001$). Its margin over the other strategies ranged from $+0.008$ (FedAdagrad) to $+0.019$ (FedCluster), and every 95\% confidence interval of the paired difference excluded zero (Holm-corrected Wilcoxon $p=0.012$, the smallest value attainable with ten seeds). AUC-PR, which concentrates on the rare deaths, separated the strategies less clearly: SCAFFOLD again had the highest mean ($0.283 \pm 0.019$), but its difference from FedAvg, $+0.004$ (95\% CI $-0.003$ to $+0.012$), was not significant, whereas its differences from the remaining five strategies were.

The eICU-CRD hospital federation produced the same ordering with smaller gaps. SCAFFOLD reached an AUC-ROC of $0.819 \pm 0.004$ and an AUC-PR of $0.418 \pm 0.009$, ranked first on both metrics in every seed (Friedman $\chi^2=50.6$ and $38.4$, $p<0.001$), and exceeded each alternative by $0.004$ to $0.009$ in AUC-ROC and by $0.011$ to $0.014$ in AUC-PR, all significant after correction. When first-day physiology was added to the eICU-CRD predictors, discrimination rose for every strategy, and SCAFFOLD remained first in all five seeds (AUC-ROC $0.878$; differences of $+0.007$ to $+0.011$ with confidence intervals excluding zero). Excluding seed 42, on which the hyperparameters had been tuned, left these results unchanged: SCAFFOLD still had the best AUC-ROC in each of the nine remaining seeds of both databases, with every confidence interval excluding zero. The control strategy showed that this advantage is not explained by SCAFFOLD's local optimizer alone. FedAvg-SGD, which shares SCAFFOLD's tuned local SGD but not its control variates, raised the AUC-ROC of FedAvg in both databases ($0.878$ against $0.874$ in MIMIC-IV and $0.816$ against $0.813$ in eICU-CRD) but not its AUC-PR in MIMIC-IV ($-0.004$, 95\% CI $-0.009$ to $0.001$), yet SCAFFOLD exceeded it (Figure~\ref{fig:paired}c) in all ten seeds of both databases, by $+0.008$ (95\% CI $0.006$ to $0.010$) in MIMIC-IV and $+0.004$ ($0.002$ to $0.005$) in eICU-CRD in AUC-ROC, and by $+0.008$ ($0.003$ to $0.014$) and $+0.004$ ($0.003$ to $0.006$) in AUC-PR (Wilcoxon $p=0.002$ in each case). In AUC-ROC, between 57\% and 63\% of SCAFFOLD's margin over FedAvg is therefore attributable to the control variates and the remainder to the change of local optimizer; in AUC-PR on MIMIC-IV, the whole margin is. The spread between the best and the worst strategy was nonetheless modest in both federations, $0.019$ in AUC-ROC for MIMIC-IV and $0.009$ for eICU-CRD, a magnitude that becomes relevant when it is compared with the effect of the protocol choices in Section~\ref{sec:protocol}.
\begin{table*}[t]\centering
\caption{Main results on the test set (mean $\pm$ SD over 10 seeds). Rank: mean Friedman rank on AUC-ROC among the seven federated strategies (1 = best). Best federated value per column in bold.}
\label{tab:main}
\footnotesize
\adjustbox{max width=\linewidth}{\begin{tabular}{@{}lcccccccc@{}}
\toprule
 & \multicolumn{4}{c}{MIMIC-IV (12 care units)} & \multicolumn{4}{c}{eICU-CRD (49 hospitals)} \\
\cmidrule(lr){2-5}\cmidrule(l){6-9}
Strategy & AUC-ROC & AUC-PR & Rank & Brier\textsuperscript{a} & AUC-ROC & AUC-PR & Rank & Brier\textsuperscript{a} \\
\midrule
FedAvg & 0.874 $\pm$ 0.004 & 0.279 $\pm$ 0.010 & 5.5 & 0.0195 & 0.813 $\pm$ 0.005 & 0.406 $\pm$ 0.008 & 3.2 & 0.0737 \\
FedProx & 0.877 $\pm$ 0.004 & 0.276 $\pm$ 0.014 & 3.8 & 0.0196 & 0.815 $\pm$ 0.005 & 0.408 $\pm$ 0.009 & 2.0 & 0.0735 \\
SCAFFOLD & \textbf{0.886 $\pm$ 0.002} & \textbf{0.283 $\pm$ 0.019} & 1.0 & \textbf{0.0194} & \textbf{0.819 $\pm$ 0.004} & \textbf{0.418 $\pm$ 0.009} & 1.0 & \textbf{0.0729} \\
FedAdagrad & 0.878 $\pm$ 0.003 & 0.277 $\pm$ 0.015 & 2.8 & 0.0196 & 0.812 $\pm$ 0.005 & 0.404 $\pm$ 0.009 & 5.2 & 0.0738 \\
FedAdam & 0.876 $\pm$ 0.004 & 0.272 $\pm$ 0.014 & 3.9 & 0.0196 & 0.810 $\pm$ 0.004 & 0.406 $\pm$ 0.008 & 6.3 & 0.0737 \\
FedYogi & 0.876 $\pm$ 0.003 & 0.272 $\pm$ 0.015 & 4.2 & 0.0196 & 0.810 $\pm$ 0.004 & 0.406 $\pm$ 0.008 & 5.7 & 0.0737 \\
FedCluster & 0.867 $\pm$ 0.005 & 0.264 $\pm$ 0.018 & 6.8 & 0.0197 & 0.812 $\pm$ 0.005 & 0.405 $\pm$ 0.009 & 4.6 & 0.0737 \\
\midrule
Centralized, no correction & 0.888 $\pm$ 0.002 & 0.294 $\pm$ 0.015 & -- & -- & 0.818 $\pm$ 0.004 & 0.417 $\pm$ 0.006 & -- & -- \\
Centralized, class weights & 0.886 $\pm$ 0.002 & 0.285 $\pm$ 0.015 & -- & -- & 0.817 $\pm$ 0.005 & 0.413 $\pm$ 0.010 & -- & -- \\
Centralized, global SMOTE & 0.838 $\pm$ 0.016 & 0.232 $\pm$ 0.022 & -- & -- & 0.769 $\pm$ 0.009 & 0.340 $\pm$ 0.013 & -- & -- \\
\bottomrule
\end{tabular}
}\\[2pt]\parbox{\linewidth}{\scriptsize \textsuperscript{a}After Platt recalibration on the validation set (Section~\ref{sec:calibration}).}
\end{table*}
\subsection{Federated versus centralized training (RQ2)}
The same experiments indicate how much of the centralized performance federated training retained. In the eICU-CRD federation, SCAFFOLD matched the centralized model trained without imbalance correction: the paired differences were $+0.001$ in AUC-ROC (95\% CI $0.000$ to $+0.003$; $p=0.11$) and $+0.001$ in AUC-PR ($-0.002$ to $+0.005$; $p=0.56$), and the same parity held with the physiological variables ($+0.001$ and $-0.003$, both intervals including zero). In the MIMIC-IV federation, SCAFFOLD fell short of that centralized model by $0.0015$ in AUC-ROC ($-0.002$ to $-0.001$) and by $0.011$ in AUC-PR ($-0.015$ to $-0.007$), both significant, but it did not differ from the centralized model trained with class weights, the treatment it shares ($+0.0005$ in AUC-ROC and $-0.002$ in AUC-PR, $p>0.2$). The centralized model with global SMOTE was the weakest of all models in both databases.
\subsection{Effect of protocol choices (RQ3)}
\label{sec:protocol}
\begin{figure*}[t]\centering
\includegraphics[width=\textwidth]{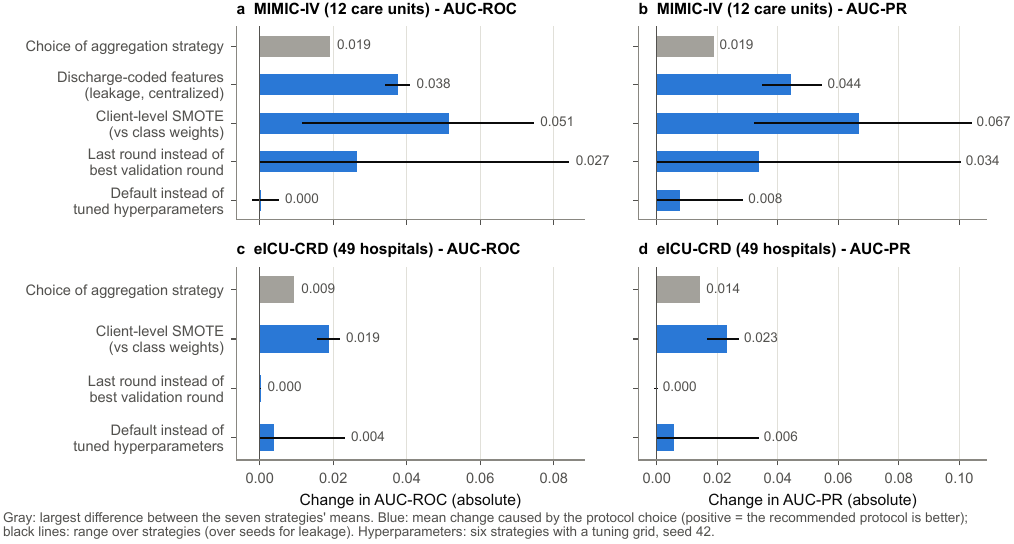}
\caption{Size of each protocol choice compared with the largest difference between aggregation strategies. Discharge-coded features were measured with a centralized gradient-boosting reference on MIMIC-IV only (5 seeds), and the effect of tuning on the tuning seed only; the other effects are means over 10 seeds.}
\label{fig:protocol-effects}
\end{figure*}

Against this background, the protocol choices changed the results by more than the choice of strategy did (Figure~\ref{fig:protocol-effects}). The discharge-coded diagnosis and procedure counts, evaluated with a centralized gradient-boosting reference on MIMIC-IV, raised AUC-ROC from $0.888$ to $0.925$ ($+0.038$) and AUC-PR from $0.317$ to $0.361$ ($+0.045$) over five seeds; the counts alone reached an AUC-ROC of $0.834$, although none of that information exists 24 hours after admission. Because this effect was measured with a different model and fewer seeds than the others, it is reported as a reference rather than compared directly with the spread between strategies; the conclusions of this section rest on the remaining choices, which were measured with the federated models themselves.

Client-level SMOTE-Tomek reduced performance across all strategies in both databases (Table~\ref{tab:imbalance}). In MIMIC-IV, where synthetic rows made up a median of 96\% and up to 99.7\% of a client's positive training cases (Table~\ref{tab:synthetic}), it reduced AUC-ROC by $0.051$ on average (from $0.012$ for SCAFFOLD to $0.074$ for FedCluster) and AUC-PR by $0.067$ (up to $0.104$ for FedAdagrad); in eICU-CRD, the reductions were about $0.019$ and $0.023$. All 14 strategy--database combinations favored class weighting over SMOTE-Tomek on both metrics, each of the 28 comparisons with the smallest attainable corrected $p$-value. Class weighting and no correction, in contrast, gave similar discrimination, with differences of at most about $0.01$ in either direction.

The choice of which round to report mattered only in the care-unit federation. Evaluating the model of the 50th round instead of the best validation round cost FedAvg $0.084$ and FedProx $0.070$ of AUC-ROC on MIMIC-IV, and less than $0.013$ for the other strategies; on eICU-CRD, the best round was, on average, one of the last, and the difference was negligible. Finally, replacing the default hyperparameters with the tuned values, a comparison available only for the tuning seed because it reuses the grid-search runs, changed little for most strategies, but it raised SCAFFOLD's AUC-PR by $0.028$ on MIMIC-IV and its AUC-ROC and AUC-PR by $0.023$ and $0.033$ on eICU-CRD, where its best local learning rate was a hundred times larger than on MIMIC-IV.
\begin{table*}[t]\centering
\caption{Class-imbalance treatment (mean of 10 seeds): class-weighted loss (Weights, the main setting), SMOTE-Tomek on each client's training data (SMOTE) and no correction (None). Brier scores before recalibration. Best value per row and metric in bold.}
\label{tab:imbalance}
\footnotesize
\adjustbox{max width=\linewidth}{\begin{tabular}{@{}lcccccccccccc@{}}
\toprule
 & \multicolumn{6}{c}{MIMIC-IV} & \multicolumn{6}{c}{eICU-CRD} \\
\cmidrule(lr){2-7}\cmidrule(l){8-13}
 & \multicolumn{3}{c}{AUC-PR} & \multicolumn{3}{c}{Brier (raw)} & \multicolumn{3}{c}{AUC-PR} & \multicolumn{3}{c}{Brier (raw)} \\
\cmidrule(lr){2-4}\cmidrule(lr){5-7}\cmidrule(lr){8-10}\cmidrule(l){11-13}
Strategy & Weights & SMOTE & None & Weights & SMOTE & None & Weights & SMOTE & None & Weights & SMOTE & None \\
\midrule
FedAvg & 0.279 & 0.228 & \textbf{0.287} & 0.071 & 0.066 & \textbf{0.019} & 0.406 & 0.384 & \textbf{0.407} & 0.167 & 0.158 & \textbf{0.074} \\
FedProx & 0.276 & 0.214 & \textbf{0.281} & 0.071 & 0.085 & \textbf{0.020} & \textbf{0.408} & 0.384 & 0.406 & 0.168 & 0.161 & \textbf{0.074} \\
SCAFFOLD & \textbf{0.283} & 0.251 & 0.272 & 0.071 & 0.136 & \textbf{0.020} & \textbf{0.418} & 0.394 & 0.417 & 0.173 & 0.166 & \textbf{0.073} \\
FedAdagrad & \textbf{0.277} & 0.173 & 0.273 & 0.075 & 0.542 & \textbf{0.020} & \textbf{0.404} & 0.387 & 0.398 & 0.167 & 0.162 & \textbf{0.074} \\
FedAdam & 0.272 & 0.191 & \textbf{0.275} & 0.073 & 0.266 & \textbf{0.020} & \textbf{0.406} & 0.379 & 0.403 & 0.172 & 0.166 & \textbf{0.074} \\
FedYogi & 0.272 & 0.192 & \textbf{0.274} & 0.074 & 0.265 & \textbf{0.020} & \textbf{0.406} & 0.379 & 0.403 & 0.171 & 0.166 & \textbf{0.074} \\
FedCluster & 0.264 & 0.206 & \textbf{0.272} & 0.059 & 0.038 & \textbf{0.020} & 0.405 & 0.383 & \textbf{0.405} & 0.167 & 0.152 & \textbf{0.074} \\
\bottomrule
\end{tabular}
}
\end{table*}
\subsection{Convergence and round selection}
\begin{figure*}[t]\centering
\includegraphics[width=\textwidth]{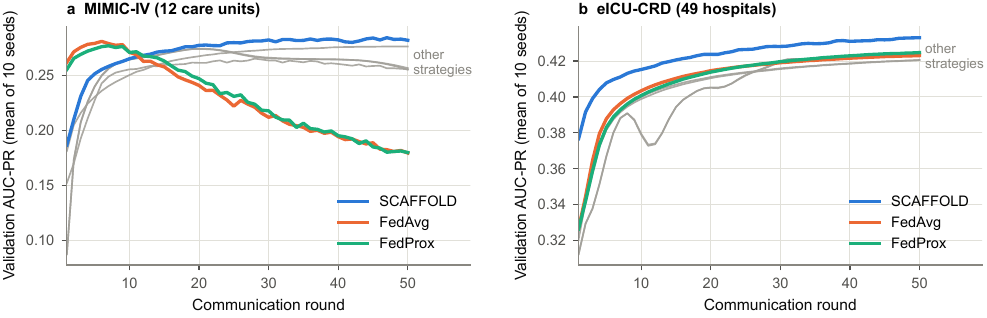}
\caption{Validation AUC-PR per communication round (mean of 10 seeds). In the care-unit federation, FedAvg and FedProx peak in the first few rounds and then degrade, whereas SCAFFOLD continues to improve.}
\label{fig:convergence}
\end{figure*}

The convergence curves in Figure~\ref{fig:convergence} show where the round effect comes from. In the MIMIC-IV federation, the validation AUC-PR of FedAvg and FedProx peaked early, at round $6.6$ and $8.2$ on average, and then declined steadily, whereas SCAFFOLD peaked at round $41$ on average and stayed near its maximum; the adaptive strategies and FedCluster fell in between. In the eICU-CRD federation, every strategy improved almost monotonically, with best rounds between $42$ and $50$, and extending training to 100 rounds on MIMIC-IV did not improve any strategy (Table~\ref{tab:robustness}).
\subsection{Calibration and recalibration}
\label{sec:calibration}
\begin{figure*}[t]\centering
\includegraphics[width=\textwidth]{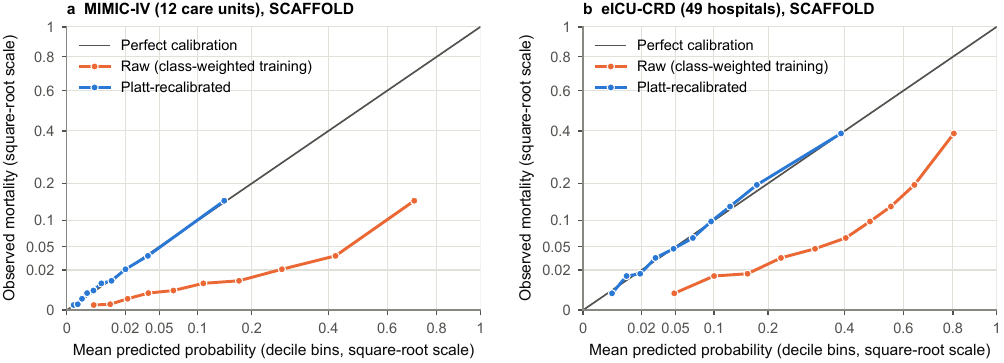}
\caption{Reliability of SCAFFOLD on the test set (seed 42, decile bins). Class-weighted training overestimates risk; Platt scaling fitted on the validation set brings the predictions onto the diagonal without changing their ranking.}
\label{fig:calibration}
\end{figure*}

The models trained with class weights discriminated well but were poorly calibrated (Table~\ref{tab:calibration} and Figure~\ref{fig:calibration}). Before recalibration, they overestimated risk in both databases, with calibration intercepts between $-2.1$ and $-3.0$, mean Brier scores of $0.071$ on MIMIC-IV and $0.169$ on eICU-CRD, and expected calibration errors of $0.15$ and $0.27$. Platt scaling on the validation set removed this bias for all seven strategies: the intercepts fell to within $0.02$ of zero, the slopes moved to between $0.98$ and $1.00$, the Brier scores dropped to $0.0196$ and $0.074$, and the calibration errors to $0.001$ and $0.006$, while AUC-ROC and AUC-PR were unchanged because the transformation is monotone. These recalibrated Brier scores equal those of the models trained without any correction ($0.019$--$0.020$ and $0.073$--$0.074$). Isotonic regression calibrated almost as well but lowered AUC-PR by about $0.013$ to $0.016$, because it maps ranges of scores onto the same value.

Miscalibration of this kind also explains the earlier observation of a strategy with a high AUC-ROC and a low F1 score. Under client-level SMOTE-Tomek, FedAdagrad on MIMIC-IV reached an AUC-ROC of $0.820$ while assigning survivors a mean predicted probability of $0.72$: at the conventional threshold of $0.5$ it classified 84\% of the test patients as deaths, with a precision of $0.029$ and a Brier score of $0.54$. The model ranked patients reasonably well, but its probabilities lacked absolute meaning, so any fixed threshold yielded a poor F1 score.
\begin{table*}[t]\centering
\caption{Calibration on the test set before $\to$ after Platt scaling (mean of 10 seeds). Intercept: calibration-in-the-large (ideal 0); slope: calibration slope (ideal 1); ECE: expected calibration error over 10 equal-frequency bins.}
\label{tab:calibration}
\footnotesize
\adjustbox{max width=\linewidth}{\begin{tabular}{@{}lcccccccc@{}}
\toprule
 & \multicolumn{4}{c}{MIMIC-IV} & \multicolumn{4}{c}{eICU-CRD} \\
\cmidrule(lr){2-5}\cmidrule(l){6-9}
Strategy & Brier & Intercept & Slope & ECE & Brier & Intercept & Slope & ECE \\
\midrule
FedAvg & 0.0709 $\to$ 0.0195 & -2.85 $\to$ 0.00 & 0.86 $\to$ 1.00 & 0.144 $\to$ 0.002 & 0.1670 $\to$ 0.0737 & -2.15 $\to$ 0.01 & 0.91 $\to$ 0.98 & 0.260 $\to$ 0.006 \\
FedProx & 0.0711 $\to$ 0.0196 & -2.83 $\to$ 0.00 & 0.88 $\to$ 1.00 & 0.144 $\to$ 0.002 & 0.1678 $\to$ 0.0735 & -2.16 $\to$ 0.01 & 0.93 $\to$ 0.98 & 0.263 $\to$ 0.006 \\
SCAFFOLD & 0.0709 $\to$ 0.0194 & -2.83 $\to$ 0.00 & 0.97 $\to$ 1.00 & 0.149 $\to$ 0.001 & 0.1726 $\to$ 0.0729 & -2.15 $\to$ 0.01 & 1.05 $\to$ 0.98 & 0.277 $\to$ 0.006 \\
FedAdagrad & 0.0747 $\to$ 0.0196 & -2.84 $\to$ 0.00 & 1.06 $\to$ 1.00 & 0.166 $\to$ 0.001 & 0.1673 $\to$ 0.0738 & -2.15 $\to$ 0.01 & 0.92 $\to$ 0.98 & 0.263 $\to$ 0.006 \\
FedAdam & 0.0730 $\to$ 0.0196 & -3.01 $\to$ 0.00 & 0.75 $\to$ 1.00 & 0.139 $\to$ 0.001 & 0.1719 $\to$ 0.0737 & -2.20 $\to$ 0.01 & 0.89 $\to$ 0.98 & 0.268 $\to$ 0.005 \\
FedYogi & 0.0737 $\to$ 0.0196 & -3.03 $\to$ 0.00 & 0.77 $\to$ 1.00 & 0.141 $\to$ 0.001 & 0.1711 $\to$ 0.0737 & -2.19 $\to$ 0.01 & 0.88 $\to$ 0.98 & 0.266 $\to$ 0.006 \\
FedCluster & 0.0591 $\to$ 0.0197 & -2.63 $\to$ 0.00 & 0.94 $\to$ 1.00 & 0.135 $\to$ 0.002 & 0.1666 $\to$ 0.0737 & -2.13 $\to$ 0.01 & 0.94 $\to$ 0.98 & 0.262 $\to$ 0.006 \\
\bottomrule
\end{tabular}
}
\end{table*}
\subsection{Robustness and per-client results}
The ranking of the strategies proved stable across the variants listed in Table~\ref{tab:experiments}. SCAFFOLD had the highest AUC-ROC in 17 of the 18 configurations in Table~\ref{tab:robustness}, including the federation variants, one- and five-local-epoch variants, partial participation by 50\% and 20\% of the clients, 100 rounds, and the logistic regression model. The single exception was the eICU-CRD federation with one extra client pooling the 158 remaining hospitals, where FedProx reached $0.819$ and SCAFFOLD $0.815$. With logistic regression, all strategies lost discrimination, and FedAdagrad lost most ($0.806$ on MIMIC-IV and $0.743$ on eICU-CRD).

Performance also varied between clients. On their own test partitions, SCAFFOLD's AUC-ROC ranged from $0.776$ (surgical ICU) to $0.893$ (medicine ward) across the MIMIC-IV care units and from $0.772$ to $0.871$ across the eICU-CRD hospitals. One epoch of local fine-tuning raised the pooled AUC-ROC of the personalized models by $0.010$ to $0.024$ for six strategies on MIMIC-IV but by only $0.003$ for SCAFFOLD, and by at most $0.002$ for any strategy on eICU-CRD. After fine-tuning on MIMIC-IV, the seven strategies fell within $0.006$ of each other ($0.885$ to $0.891$), and personalized FedCluster slightly exceeded personalized SCAFFOLD ($0.891$ against $0.889$).

Performance was similar between sexes (Table~\ref{tab:subgroups}), but it was lower for patients aged 65 years or older ($0.837$ against $0.908$ on MIMIC-IV and $0.788$ against $0.839$ on eICU-CRD for SCAFFOLD), a common pattern when age itself is a strong predictor, and it varied across racial and ethnic groups, from $0.847$ to $0.910$ on MIMIC-IV and from $0.800$ to $0.846$ on eICU-CRD. SCAFFOLD's AUC-ROC was at least as high as FedAvg's in every subgroup across both databases, so its advantage was not achieved at the expense of any group. Taken together, these results describe a consistent but small advantage of one aggregation strategy against much larger effects of the evaluation protocol, a contrast that the next section interprets.

\section{Discussion}
\label{sec:discussion}
This study asked which aggregation strategy performs best when the evaluation protocol is controlled, how close federated training comes to centralized training, and how much the protocol itself changes the conclusions. The answers form a consistent picture. SCAFFOLD was the best strategy in both federations and in almost every variant, but its advantage over the alternatives was small, ranging from $0.004$ to $0.019$ in AUC-ROC. On the federation of 49 hospitals, federated training matched centralized training. By contrast, three common protocol choices, namely predictors recorded at discharge, client-level oversampling, and the evaluation of the last round, each moved discrimination in the care-unit federation by more than the whole range between the strategies, and oversampling did so in the hospital federation too. The remainder of this section interprets these findings, relates them to previous work, and draws practical recommendations from them.

The advantage of SCAFFOLD is consistent with the problem it was designed to solve. When clients hold different data, each local model drifts towards the optimum of its own data, and averaging these drifted models does not move the global model towards the optimum of the federation~\cite{karimireddy2020}. SCAFFOLD subtracts an estimate of each client's drift from every local gradient, whereas FedProx only limits how far a local model may move, not the direction in which it moves~\cite{li2020fedprox}. The convergence curves support this interpretation: in the care-unit federation, where a single client holds most of the data, FedAvg and FedProx reached their peak validation performance within the first 10 rounds and then declined, whereas SCAFFOLD continued to improve until late in training. The correction comes with a practical condition, however. SCAFFOLD collapsed when it was combined with the adaptive local optimizer used by the other strategies, and its best local learning rate differed a hundredfold between the two databases. The FedAvg-SGD control separates these two ingredients: plain local SGD with a tuned learning rate recovered about two fifths of SCAFFOLD's margin over FedAvg in AUC-ROC and none of it in AUC-PR on MIMIC-IV, so the drift correction, not the change of optimizer, accounts for most of the advantage. A practitioner who adopts SCAFFOLD, therefore, has to tune it with plain local SGD, rather than reuse the settings of another method. Its advantage also concerns the global model shared by all clients: when each care unit fine-tuned the global model on its own data, the other strategies caught up, which suggests that in a strongly unbalanced federation local adaptation can compensate for a weaker aggregation rule.

The structure of the federation shaped how much the choice of strategy mattered. In the MIMIC-IV federation, the Emergency Department accounted for 78.7\% of the stays, and in that setting, the strategies differed more ($0.019$ in AUC-ROC versus $0.009$ in eICU-CRD), and the round at which the model was taken changed the results by up to $0.084$. In the eICU-CRD federation, where no hospital accounted for more than 5.5\% of the stays, the strategies behaved similarly, and all of them continued to improve until round 40 or later. This contrast suggests that the dominance of one client, rather than heterogeneity alone, is what makes the aggregation rule matter. Previous studies that partitioned MIMIC-III at random found federated performance close to centralized performance~\cite{lee2020fl,mondrejevski2022}, but random partitions produce clients with similar data; the care-unit federation shows that a dominant and distinct client is what makes the strategy and the evaluated round matter. It also qualifies a previous explanation for the weak results of FedProx, namely that the eICU-CRD hospitals are too homogeneous for its proximal term to help~\cite{dang2022}: in the strongly heterogeneous care-unit federation, FedProx did not exceed FedAvg in AUC-PR either ($0.276$ against $0.279$). Its benefit appeared as stability, a slower degradation after the early peak, rather than as a higher maximum. Because the same pipeline and predictors were used in both federations, these differences can be attributed to the structure of the federations rather than to data preparation.

These results agree only in part with the closest previous comparison~\cite{dang2022}. That study also found small differences between aggregation strategies on eICU-CRD and a best federated model $0.012$ AUC-ROC below the centralized one ($0.906$ against $0.918$), but it ranked FedAvg and FedAvgM first and FedProx last, did not evaluate SCAFFOLD, used shared hyperparameters whose values were not reported, and compared a single run of each method. Its protocol also used the federated models from the last round and the centralized model from the best validation epoch. The present results show that such an asymmetry is not neutral: had the last round been reported here, FedAvg would have lost $0.084$ of AUC-ROC on MIMIC-IV, enough to create or hide a gap between federated and centralized training. Selecting the round on the test data, as one of the studies in Table~\ref{tab:related} did~\cite{kerkouche2021}, introduces the opposite bias, because the reported performance is then the maximum over many evaluations of the same test set. The absolute AUC-ROC in the present study is lower than in that work ($0.819$ against about $0.90$ on eICU-CRD) because its 82 predictors included vital signs and the Glasgow Coma Scale; when comparable first-day physiology was added, AUC-ROC rose to $0.878$, and the ranking of the strategies and the parity with centralized training were unchanged.

The effects of the protocol choices extend findings from centralized clinical prediction to the federated setting. Leakage from predictors unavailable at prediction time is a known source of overoptimism in machine learning~\cite{kapoor2023}, and here, a pair of discharge-coded counts added almost $0.04$ to AUC-ROC. Resampling is known to distort predicted risks without improving discrimination in centralized models~\cite{goorbergh2022}; in the federated setting, client-level SMOTE-Tomek also reduced discrimination across all strategies and both databases. A plausible explanation is that the synthetic cases, which accounted for up to 99.7\% of a client's positives, reproduced the few real deaths for each client and the local idiosyncrasies of those cases, and that aggregation then propagated these artifacts into the global model. Class weighting avoided this loss but still distorted the predicted risks, and Platt scaling on validation predictions brought them back to the calibration of models trained without any correction. Because Platt scaling fits only two parameters, a deployment can estimate it by averaging the clients' gradients of a two-parameter logistic regression, without sharing any record, so that the risk estimates delivered to clinicians need not depend on the imbalance treatment chosen for training. Selecting the best round by validation AUC-PR is less direct, because AUC-PR cannot be computed from per-client summaries; it requires the clients' validation scores, or histograms of them, to be pooled through secure aggregation.

Four practical recommendations for studies of federated learning on electronic health records follow from these observations. First, predictors should be restricted to information available at the moment of prediction, and features derived from billing or discharge codes should be excluded or justified. Second, class imbalance is better handled by weighting the loss, or by no correction at all, followed by recalibration, than by oversampling each client. Third, every method, whether federated or centralized, should be selected by the same validation rule, and its hyperparameters should be tuned on validation data rather than being fixed in common. Fourth, comparisons should be repeated across several seeds and tested using procedures designed for multiple methods~\cite{demsar2006}, reporting both calibration and discrimination. Under these conditions, the choice of aggregation strategy still matters, but by less than the protocol does.
\subsection{Limitations}
The study has limitations that bound these conclusions. The predictors were restricted to demographics and a laboratory panel available in both databases, which made the two federations comparable but kept discrimination below that of models with richer clinical data; the sensitivity analysis with physiological variables mitigates this concern for eICU-CRD only. Both databases are from the United States, and eICU-CRD covers 2014 and 2015, so the conclusions may not generalize to hospitals with different populations, practices, or record systems, including those in the Brazilian public health system. The federations were simulated on a single machine: communication costs, client failures, and the secure aggregation of shared preprocessing statistics and validation scores (used for round selection and recalibration) were not modeled, and no formal privacy guarantee, such as differential privacy, was applied. The same pooled validation set served three purposes, tuning, round selection and recalibration; the test set was never used for any of them, but the accumulated decisions may make validation performance slightly optimistic. The hyperparameters were tuned on a single seed of the main configuration and reused across the variants, which may explain the single variant in which SCAFFOLD was not the best. The tuning budget also differed between strategies, from one configuration for FedAvg to 14 for SCAFFOLD; the FedAvg-SGD control equalizes the budget for the local optimizer only, and larger grids for the adaptive strategies or FedCluster might narrow their gap to SCAFFOLD. Only a multilayer perceptron and a logistic regression were evaluated, and personalization was limited to local fine-tuning. The degree of heterogeneity between clients was not varied synthetically, for example with Dirichlet partitions; the two natural federations, one dominated by a single client and one balanced, take that role. Fairness was examined only descriptively, by comparing discrimination across subgroups, and some groups, such as Asian patients in eICU-CRD, contributed few deaths. Finally, the MIMIC-IV federation partitions a single hospital by care unit, and some of its clients, such as the Discharge Lounge, are administrative locations rather than clinical units; it represents an extreme intra-hospital imbalance by design and is not a substitute for a federation of independent institutions, which is the role of the eICU-CRD federation in this study.

\section{Conclusion}
\label{sec:conclusion}
This study compared seven federated aggregation strategies for in-hospital mortality prediction under a protocol that treated all strategies equally, on an intra-hospital and an inter-hospital federation built with the same pipeline. Regarding the first research question, SCAFFOLD was the most accurate strategy in both federations and almost every variant, provided it trained locally with gradient descent at a tuned learning rate, and a control with the same local optimizer showed that its advantage came mainly from its drift correction; that advantage was consistent but small, and larger when a single client dominated the federation. Regarding the second question, federated training matched centralized training across a federation of 49 hospitals, and Platt scaling on validation predictions repaired the risk estimates that class weighting distorts, at no cost to discrimination. Regarding the third question, the protocol mattered more than the algorithm: in the care-unit federation, predictors recorded at discharge, oversampling inside each client, and the evaluation of the last training round each changed discrimination by more than the whole spread between strategies; oversampling did so in the hospital federation as well, and tuning was decisive for SCAFFOLD.

For practitioners, the results imply that the evaluation protocol deserves as much attention as the choice of aggregation strategy, and that a reported ranking of strategies is only as reliable as the protocol behind it. Future work should evaluate these strategies in real deployments, with communication constraints, secure aggregation of shared statistics, and formal privacy guarantees such as differential privacy; extend the comparison to data from other health systems, including the Brazilian public health system; and include other model families, such as gradient-boosted trees, and personalized federated learning.

\section*{Data and code availability}
MIMIC-IV (version 3.1) and eICU-CRD (version 2.0) are available from PhysioNet to credentialed users who complete the required training and sign the corresponding data use agreements~\cite{mimiciv31,eicu20}; under these agreements, the derived cohorts cannot be redistributed. The code that builds the cohorts from the PhysioNet files, runs all experiments, and produces all figures and tables is available in a public mirror of the study repository (\url{https://anonymous.4open.science/r/protocol-choices-federated-mortality-D18F/}).
\section*{Ethics}
This retrospective study used only de-identified data. The collection and sharing of both databases were approved by the institutional review boards of the contributing institutions with a waiver of informed consent, as described in their original publications~\cite{johnson2023,pollard2018}, and access was obtained under the PhysioNet data use agreements.
\section*{Patient and public involvement}
Patients and the public were not involved in the design, conduct or reporting of this secondary analysis of public, de-identified data.
\section*{Funding}
This research received no specific funding.
\section*{Acknowledgements}
The authors would like to acknowledge the support of the Software Engineering and Automation Research Laboratory, where the research was developed and conducted. The infrastructure and resources provided were crucial for completing the work.
\section*{Conflicts of interest}
The authors declare no conflicts of interest.
\section*{Declaration of AI-Assisted Use in Writing}
Generative artificial intelligence tools were used to revise the manuscript's language and to review and refactor code. The authors reviewed all suggestions and take full responsibility for the content of the article. No such tool was used to generate, alter, or analyze data, or to create or modify figures.

\bibliographystyle{unsrtnat}
\bibliography{refs}

\clearpage
\appendix
\section{Hyperparameters, subgroups, oversampling and robustness}
\begin{table}[h!]\centering
\caption{Per-strategy tuning grids (validation AUC-PR, seed 42) and selected values. Configs: number of configurations tried on MIMIC-IV/eICU-CRD. FedAvg has no strategy-specific hyperparameter; FedAvg-SGD is the control for SCAFFOLD's local optimizer. Shared settings: Adam with learning rate $10^{-3}$ for local training (SGD for SCAFFOLD), batch 512, 3 local epochs, 50 rounds.}
\label{tab:tuning}
\footnotesize
\adjustbox{max width=\linewidth}{\begin{tabular}{@{}lllll@{}}
\toprule
Strategy & Grid & Configs & Selected, MIMIC-IV & Selected, eICU-CRD \\
\midrule
FedProx & $\mu \in \{0.001, 0.01, 0.1, 1\}$ & 4/4 & $\mu$ = 0.001 & $\mu$ = 0.1 \\
SCAFFOLD & local SGD lr $\in \{0.003, 0.01, 0.03, 0.1, 0.3, 1, 3^{\mathrm{b}}\}$\textsuperscript{a}, $\eta_g \in \{0.5, 1\}$ & 12/14 & $\eta_g$ = 1, lr = 0.01 & $\eta_g$ = 1, lr = 1 \\
FedAdagrad & $\eta \in \{0.001, 0.01, 0.1\}$, $\tau \in \{10^{-3}, 10^{-2}\}$ & 6/6 & $\eta$ = 0.01, $\tau$ = 0.001 & $\eta$ = 0.01, $\tau$ = 0.001 \\
FedAdam & $\eta \in \{0.001, 0.01, 0.1\}$, $\tau \in \{10^{-3}, 10^{-2}\}$ & 6/6 & $\eta$ = 0.01, $\tau$ = 0.01 & $\eta$ = 0.01, $\tau$ = 0.001 \\
FedYogi & $\eta \in \{0.001, 0.01, 0.1\}$, $\tau \in \{10^{-3}, 10^{-2}\}$ & 6/6 & $\eta$ = 0.01, $\tau$ = 0.01 & $\eta$ = 0.01, $\tau$ = 0.001 \\
FedCluster & DBSCAN radius quantile $\in \{0.25, 0.5, 0.75\}$ & 3/3 & q = 0.75 & q = 0.25 \\
\midrule
FedAvg-SGD (control) & local SGD lr $\in \{0.003, 0.01, 0.03, 0.1, 0.3, 1, 3^{\mathrm{b}}\}$\textsuperscript{a} & 6/7 & lr = 0.1 & lr = 1 \\
\bottomrule
\end{tabular}
}\\[2pt]
\parbox{\linewidth}{\scriptsize \textsuperscript{a}1 and 3 were added after the optimum landed on the largest value of the initial grid. \textsuperscript{b}eICU-CRD only.}
\end{table}
\begin{table}[h!]\centering
\caption{Test AUC-ROC by sociodemographic subgroup for SCAFFOLD and FedAvg (main experiment, mean of 10 seeds). Deaths: mean number of deaths in the test set per seed.}
\label{tab:subgroups}
\footnotesize
\adjustbox{max width=0.7\linewidth}{\begin{tabular}{@{}lcccccc@{}}
\toprule
 & \multicolumn{3}{c}{MIMIC-IV} & \multicolumn{3}{c}{eICU-CRD} \\
\cmidrule(lr){2-4}\cmidrule(l){5-7}
Subgroup & Deaths & SCAFFOLD & FedAvg & Deaths & SCAFFOLD & FedAvg \\
\midrule
Female & 1031 & 0.887 & 0.876 & 934 & 0.816 & 0.810 \\
Male & 1204 & 0.884 & 0.871 & 1060 & 0.822 & 0.815 \\
\midrule
Under 65 years & 780 & 0.908 & 0.894 & 684 & 0.839 & 0.833 \\
65 years or older & 1455 & 0.837 & 0.821 & 1311 & 0.788 & 0.781 \\
\midrule
White & 1413 & 0.871 & 0.860 & 1545 & 0.818 & 0.815 \\
Black & 222 & 0.882 & 0.867 & 225 & 0.815 & 0.807 \\
Hispanic & 66 & 0.906 & 0.853 & 79 & 0.828 & 0.817 \\
Asian & 85 & 0.847 & 0.831 & 35 & 0.800 & 0.791 \\
Other or unknown & 450 & 0.910 & 0.901 & 110 & 0.846 & 0.837 \\
\bottomrule
\end{tabular}
}
\end{table}
\begin{table}[h!]\centering
\caption{Training data of each client under client-level SMOTE-Tomek (mean over 10 seeds): real deaths in the training partition, synthetic deaths added by SMOTE, and the share of synthetic cases among the client's positive training cases.}
\label{tab:synthetic}
\footnotesize
\adjustbox{max width=0.75\linewidth}{\begin{tabular}{@{}lrrrr@{}}
\toprule
Client & Training stays & Real deaths & Synthetic deaths & Synthetic share \\
\midrule
Emergency Department & 261,899 & 6,008 & 249,804 & 97.7\% \\
Discharge Lounge & 34,586 & 119 & 34,347 & 99.7\% \\
Medicine/Cardiology & 7,840 & 104 & 7,632 & 98.7\% \\
Medicine & 7,805 & 230 & 7,336 & 97.0\% \\
Hematology/Oncology & 7,039 & 213 & 6,605 & 96.9\% \\
Transplant & 4,176 & 152 & 3,870 & 96.2\% \\
Hematology/Oncology Intermediate & 3,588 & 138 & 3,310 & 96.0\% \\
Coronary Care Unit (CCU) & 1,832 & 225 & 1,381 & 86.0\% \\
Medical Intensive Care Unit (MICU) & 1,479 & 305 & 865 & 73.9\% \\
Medical/Surgical Intensive Care Unit (MICU/SICU) & 1,112 & 175 & 759 & 81.2\% \\
Cardiac Vascular Intensive Care Unit (CVICU) & 828 & 71 & 684 & 90.5\% \\
Surgical Intensive Care Unit (SICU) & 610 & 81 & 447 & 84.6\% \\
\midrule
eICU-CRD, 49 hospitals (median) & 1,168 & 105 & 951 & 89.0\% \\
eICU-CRD, 49 hospitals (range) & 572--3,830 & 69--356 & 372--3,241 & 79.0--94.3\% \\
\bottomrule
\end{tabular}
}
\end{table}
\begin{table}[h!]\centering
\caption{Robustness: test AUC-ROC (mean over the number of seeds in parentheses) of every strategy in every variant. Best per row in bold.}
\label{tab:robustness}
\footnotesize
\adjustbox{max width=\linewidth}{\begin{tabular}{@{}lccccccc@{}}
\toprule
Variant (seeds) & FedAvg & FedProx & SCAFFOLD & FedAdagrad & FedAdam & FedYogi & FedCluster \\
\midrule
MIMIC-IV, main (10) & 0.874 & 0.877 & \textbf{0.886} & 0.878 & 0.876 & 0.876 & 0.867 \\
+ pooled other units (5) & 0.882 & 0.885 & \textbf{0.892} & 0.887 & 0.884 & 0.883 & 0.876 \\
Emergency Dept.\ split in 4 (5) & 0.876 & 0.877 & \textbf{0.886} & 0.880 & 0.876 & 0.876 & 0.875 \\
Five largest units (5) & 0.882 & 0.885 & \textbf{0.889} & 0.886 & 0.882 & 0.883 & 0.875 \\
1 local epoch (3) & 0.878 & 0.876 & \textbf{0.886} & 0.878 & 0.877 & 0.877 & 0.874 \\
5 local epochs (3) & 0.876 & 0.875 & \textbf{0.883} & 0.876 & 0.874 & 0.874 & 0.856 \\
50\% participation (3) & 0.877 & 0.878 & \textbf{0.886} & 0.877 & 0.875 & 0.875 & 0.874 \\
100 rounds (1) & 0.879 & 0.878 & \textbf{0.884} & 0.879 & 0.879 & 0.879 & 0.865 \\
Logistic regression (3) & 0.855 & 0.854 & \textbf{0.861} & 0.806 & 0.852 & 0.851 & 0.856 \\
\midrule
eICU-CRD, main (10) & 0.813 & 0.815 & \textbf{0.819} & 0.812 & 0.810 & 0.810 & 0.812 \\
+ pooled other hospitals (5) & 0.814 & \textbf{0.819} & 0.815 & 0.815 & 0.812 & 0.812 & 0.817 \\
Hospitals with $\geq$200 deaths (5) & 0.807 & 0.810 & \textbf{0.814} & 0.808 & 0.804 & 0.805 & 0.809 \\
1 local epoch (3) & 0.806 & 0.806 & \textbf{0.813} & 0.808 & 0.809 & 0.809 & 0.804 \\
5 local epochs (3) & 0.809 & 0.813 & \textbf{0.816} & 0.810 & 0.808 & 0.808 & 0.810 \\
50\% participation (3) & 0.810 & 0.813 & \textbf{0.817} & 0.810 & 0.809 & 0.809 & 0.810 \\
20\% participation (3) & 0.810 & 0.812 & \textbf{0.813} & 0.809 & 0.809 & 0.809 & 0.810 \\
Logistic regression (3) & 0.778 & 0.778 & \textbf{0.787} & 0.743 & 0.782 & 0.782 & 0.776 \\
\midrule
eICU-CRD + day-1 physiology (5) & 0.869 & 0.871 & \textbf{0.878} & 0.869 & 0.867 & 0.867 & 0.869 \\
\bottomrule
\end{tabular}
}
\end{table}

\end{document}